\pdfoutput=1

\documentclass[11pt]{article}
\usepackage[hyperref]{acl2021}
\usepackage[T1]{fontenc}
\usepackage{times}
\usepackage[utf8]{inputenc} 
\usepackage{amsfonts}
\usepackage{microtype} 
\usepackage{inconsolata}
\usepackage{multirow}  
\usepackage{booktabs}

\usepackage{latexsym}
\usepackage{graphicx}
\graphicspath{{media/}}
\usepackage{amsmath}
\usepackage{enumitem}
\usepackage{siunitx}

\renewcommand{\newlinechar}{\verb~\textless newline\textgreater~}
\newcommand{\avg}[2]{\num{#1} ($\pm$ \num{#2})}

\usepackage{todonotes} 

\aclfinalcopy 
\def\aclpaperid{1} 
\title{Decoding Methods for Neural Narrative Generation}

\author{
Alexandra DeLucia\thanks{$^*$Equal contribution.}
\and
Aaron Mueller$^*$\\
Johns Hopkins University\\
{\{\texttt{aadelucia}, \texttt{amueller}\}\texttt{@jhu.edu}}
\And
Xiang Lisa Li\thanks{$^\dagger$Work performed while at Johns Hopkins University.} \\ Stanford University\\\texttt{xlisali@stanford.edu}
\AND
Jo\~ao Sedoc$^\dagger$\\ 
New York University\\\texttt{jsedoc@stern.nyu.edu}
}

\begin{document}
\maketitle

\begin{abstract}
    Narrative generation is an open-ended NLP task in which a model generates a story given a prompt. The task is similar to neural response generation for chatbots; however, innovations in response generation are often not applied to narrative generation, despite the similarity between these tasks. We aim to bridge this gap by applying and evaluating advances in decoding methods for neural response generation to neural narrative generation. In particular, we employ GPT-2 and perform ablations across nucleus sampling thresholds and diverse decoding hyperparameters---specifically, maximum mutual information---analyzing results over multiple criteria with automatic and human evaluation. We find that (1) nucleus sampling is generally best with thresholds between 0.7 and 0.9; (2) a maximum mutual information objective can improve the quality of generated stories; and (3) established automatic metrics do not correlate well with human judgments of narrative quality on any qualitative metric.
\end{abstract}

\section{Introduction}

Narrative generation (or story generation) is the task of generating a creative response given an input prompt. This output can be a story closure, a paragraph, or a structured story with multiple paragraphs. This input and output setup is similar to the response generation task of chatbots, as both tasks convert some variable-length sequential input from a user to an automatically generated variable-length sequential output. Thus, the neural models and methods proposed to date for story generation and dialogue generation have been similar.

\begin{figure}
    \centering
    \includegraphics[width=0.47\textwidth]{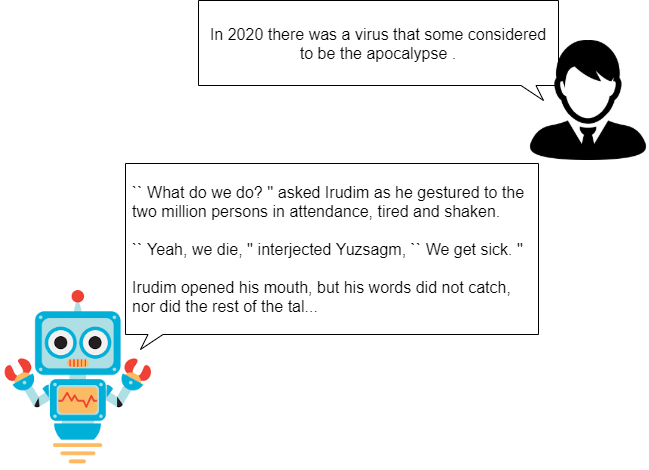}
    \caption{Example of interactive narrative generation. A user provides a prompt to our model (fine-tuned GPT-2 model), and the model responds with a story conditioned on the prompt.} 
    \label{fig:narrative-gen-ex}
\end{figure}

However, as narrative generation is largely focused on coherence across long outputs, the strategies used in this subfield have evolved separately from those in chatbot response generation; the latter has been more concerned with generating interesting and diverse---and typically \emph{short}---outputs. Thus, while many beneficial techniques may have arisen from one domain, they are not often employed in the other. One decoding method, nucleus sampling~\citep{Holtzman2020}, has recently been applied to narrative generation \citep{ippolito2020}, but a thorough evaluation of its various $p$ thresholds has not been performed with human judgments using narrative-specific criteria, as this can be time- and labor-intensive. Also, recent advances in decoding methods for response generation---notably, the application of the maximum mutual information (MMI) objective \citep{li_diversity-promoting_2016}---have resulted in more interesting dialog according to human evaluators \citep{zhang2019dialogpt}; nonetheless, this also has not been applied to narrative generation. Indeed, the MMI objective has been confined to short-form and less open-ended generation tasks thus far.

Thus, we apply techniques from neural response generation to neural narrative generation in order to investigate the potential benefits---and pitfalls---of applying these methods in this underexplored domain. This study aims to connect research developments across tasks by sweeping various thresholds of nucleus sampling and the application of diverse decoding to generate more long-form creative outputs. We perform human and automatic evaluations of automatically generated stories in these settings in order to investigate the following phenomena:

\begin{enumerate}[noitemsep,topsep=0pt, partopsep=0pt]
    \item The effect of the nucleus sampling threshold $p$ on narrative quality.
    \item The effect of the maximum mutual information (MMI, \citealt{jiweili16}) diverse decoding objective with various diversity strengths $\lambda$ on narrative quality.
    \item The correlation (or lack thereof) between human evaluations of narrative quality and automatic metrics for response generation.
\end{enumerate}

As this domain generates longer and less constrained outputs than other natural language generation (NLG) tasks, we expect to find different ideal settings than those found for short-form or constrained generation.

Our preprocessing, training, generation, and analysis scripts are available publicly.\footnote{\url{https://github.com/AADeLucia/gpt2-narrative-decoding}}

\section{Related Work}\label{sec:related}
    \paragraph{Narrative generation tasks} Work on narrative generation is split between cloze tasks, open-ended generation, and guided generation. In a cloze task, a full story except for a final word, phrase, or sentence is given, and a model generates a completion. This could be cast as a short generation problem---or, more commonly in this domain, a multiple-choice problem \citep{mostafazadeh-etal-2016-corpus,weston2015aicomplete,hill2015goldilocks,ippolito2019unsupervised}.
    
    Open-ended generation is the task of generating long-form output conditioned on a prompt (Figure~\ref{fig:narrative-gen-ex}).
    \citet{fan_hierarchical_2018} create a paired prompt and response dataset from the subreddit {\tt r/WritingPrompts}\footnote{\url{https://www.reddit.com/r/WritingPrompts/}} to train a sequence-to-sequence ``fusion model.''
    \citet{see-etal-2019-massively} extend \citet{fan_hierarchical_2018}, but use GPT-2 small and perform a top-$k$ decoding parameter sweep. 
    We focus on this open-ended narrative generation task in our investigation, but primarily focus on GPT-2 Medium and on the effect of nucleus sampling thresholds and diverse decoding strengths on narrative quality. While \citet{nadeem2020systematic} similarly perform a hyperparameter search over sampling algorithms in a language generation setting, they perform human evaluations using a convincingness metric on a short-form news generation task; long-form narrative generation is not bound by realism (and may actually benefit from less realistic output), and thus requires different metrics and evaluation setups.
    
    Guided generation is the middle ground of cloze and open-ended generation. The model is provided more context, such as characters, plot information, and potentially other information, and then generates a story based on all of the provided structural and semantic information \citep{peng_towards_2018,akoury_storium_2020}.

    \begin{table}[]
        \centering
        \begin{tabular}{l}
        \toprule
        \begin{minipage}{0.95\linewidth}
        [ WP ] You live in a world where there has never been sickness , and you are the first to have ever experienced being sick .
        \end{minipage} \\
        \toprule
        \begin{minipage}{0.95\linewidth}
        I open my eyes in a panic , sweat beading and then falling down my face . I look around and the sun in shining through the maroon curtains of my studio apartment . Everything seems to be as I left it the afternoon before , but there is a heavy , unfamiliar air in the room .
        \end{minipage} \\
        \bottomrule
        \end{tabular}
        \caption{Example prompt and response excerpt from \textsc{WritingPrompts}.}
        \label{tab:wp-dataset-ex}
    \end{table}
    
    \paragraph{Decoding methods for generation} Decoding refers to the inference methods used in natural language generation; given input sequence $S$, how should we construct the output sequence $T$? Since finding the exact most probable token at each time step often does not produce human-like or high-quality results \citep{zhang2020trading,Holtzman2020}, search and sampling are used to overcome label bias and generate more human-like language. One popular search method is beam search, where at each time step, the algorithm keeps track of the top $B$ most probable partial hypotheses. When $B=1$, this method reduces to the \emph{greedy} decoder, which chooses the argmax over the model's token distribution at each time step. 
    
    An alternative to search is sampling-based approaches, which select a token with likelihood proportional to a (typically constrained) probability distribution at each time step. Such methods include top-$k$ \citep{fan_hierarchical_2018} which restricts the sampling space to the top $k$ most probable tokens at every time step, and ``nucleus sampling''\footnote{Also referred to as ``top-$p$''.} \citep{Holtzman2020} which thresholds the cumulative token probability distribution according to a hyperparameter $p$. We focus on nucleus sampling, as it has tended to be a more effective decoding method in various response generation settings \citep{zhang2020trading,ippolito2020}.
    
    An approach to control sampling is temperature \citep{ackley1985temperature}, which modifies the softmax estimating the token probability distribution. This has been applied widely in neural text generation \citep{ficler2017temperature,caccia2018language}, especially when using top-$k$ or random sampling. Low temperatures bias the model toward high-probability events, which tends to increase generation quality while decreasing token diversity \citep{hashimoto2019unifying}. Temperature sampling has been investigated extensively in natural language generation over multiple sampling methods, and nucleus sampling has been found to be a more effective method of controlling the sampling distribution \citep{Holtzman2020}, so we do not investigate this here.

    \paragraph{Decoding objective} In chatbot response generation, top-$k$ and nucleus sampling have been known to generate fluent, but uninteresting and simple high-probability responses which do not address the input \citep{jiweili16}. This issue is commonly referred to as the ``I don't know'' problem, where the response to all inputs is often the high-probability phrase ``I don't know.'' Proposed solutions to this response blandness issue involve altering the decoding objective. Some recent work in this domain includes \citet{RyoNakamura18}, who use Inverse Token Frequency to reweight generated tokens. \citet{xu18} and \citet{zhang18} use adversarial loss to optimize for diversity, informativeness, and fluency. \citet{martins2020sparse} propose entmax sampling to generate more effectively from sparse distributions and address the train-test mismatch in text generation.
    
    Another approach explores variants of the standard log-likelihood loss, applying different objectives during inference. An example of this is maximum mutual information (MMI, \citealt{jiweili16}), an objective that promotes more diverse responses in the neural response generation task. This mitigates the ``I don't know'' problem in which all responses tend to converge to some high-probability sequence with no real content conveyed in response to the input sequence. Two versions are introduced in \citet{jiweili16}: bidirectional (MMI-bidi) and an anti-language model (MMI-antiLM) objective. The typical decoding objective is defined as
    \begin{equation*}
        \hat{T} = \arg\max_T \log p(T \mid S)
    \end{equation*}
    where $S$ is the input sequence, $T$ is a possible target sequence, and $\hat{T}$ is the selected target. We use a slightly modified form of the MMI-antiLM objective \citep{li_diversity-promoting_2016}, defined as follows:
    \begin{equation*}
        \hat{T} = \arg\max_T \log p(T\mid S) - \lambda \log p(T)
    \end{equation*}
    where $\lambda$ is a hyperparameter controlling the degree to which the language modeling objective is subtracted from the sequential transduction objective. Intuitively, this is meant to increase the likelihood of relevant targets while penalizing popular generic responses (e.g. ``okay'').
    
    This diverse decoding objective has been applied to response generation but has not yet been applied to the narrative generation task; here, we evaluate the effect of the MMI-antiLM objective on narrative generation quality.

\section{Experimental Setup}

\subsection{Dataset}\label{sec:data}

For our task of narrative generation, we train on \citet{fan_hierarchical_2018}'s long-form response dataset \textsc{WritingPrompts}.\footnote{\url{https://github.com/pytorch/fairseq/blob/master/examples/stories/README.md}} This dataset was built from the subreddit \texttt{r/WritingPrompts}\footnote{\url{https://www.reddit.com/r/WritingPrompts/}}, where users post a ``prompt" consisting of up to a few sentences, and other users reply to the post with a story continuing the prompt (the ``response"). An example prompt and response pair is in Table~\ref{tab:wp-dataset-ex}. 

To create datasets of varying lengths---and to make the dataset compatible with our model (GPT-2, discussed more in \S\ref{sec:model})---we preprocess the \textsc{WritingPrompts} dataset as follows:
\begin{enumerate}[noitemsep,topsep=0pt, partopsep=0pt]
    \item Remove all prompts that are not tagged with \verb~[ WP ]~. Other tags in {\tt r/WritingPrompts} have response requirements and constraints, such as having to occur in an established universe or not including particular tokens; we want only unconstrained responses.
    \item\label{step:length} Create different versions of each response by using all content from (1) before the first line break/the first 100 tokens, (2) before the third line break/the first 256 tokens, and (3) the entire response/the first 1024 tokens, respectively. These are referred to as the ``small'', ``medium'', and ``large'' datasets/response lengths, and are treated as separate corpora. Thus, we have 3 train, validation, and test corpora for a total of 9.
    \item Combine the source (prompt) and target (response) strings into one, as in Figure~\ref{fig:preprocessing}. 
\end{enumerate}

\begin{figure}
    \centering
    \small{
    \verb~<|startoftext|>~ \verb~[WP]~ PROMPT \verb~[RESPONSE]~ RESPONSE \verb~<|endoftext|>~}
    \caption{Each prompt/response pair from \textsc{WritingPrompts} was formatted for compatibility with GPT-2. Note: ``[WP]" and ``[RESPONSE]" are defined as special tokens so that they are not split into subword units.}
    \label{fig:preprocessing}
\end{figure}

\begin{table}[]
    \centering
    \resizebox{1\linewidth}{!}{
    \begin{tabular}{llcr}
    \toprule
        \textbf{Fold} & \textbf{Size} & \textbf{Tokens Per Example} & \textbf{Total Tokens} \\
    \midrule
        \multirow{3}{*}{Train} & Small & \avg{92.9}{82.8} & $21.4$M \\
         & Medium & \avg{206.0}{128.2} & $47.5$M \\
         & Large & \avg{718.4}{458.9} & $165.8$M \\
         \midrule
        \multirow{3}{*}{Valid} & Small & \avg{92.9}{80.2} & $1.2$M \\
        & Medium & \avg{206.1}{128.3} & $2.8$M \\
        & Large & \avg{714.4}{463.3} & $9.5$M \\
        \midrule
        \multirow{3}{*}{Test} & Small & \avg{91.4}{79.4} & $1.2$M \\
        & Medium & \avg{204.7}{124.1} & $2.6$M \\
        & Large & \avg{720.4}{455.9} & $9.3$M\\ 
    \bottomrule
    \end{tabular}}
    \caption{Corpus sizes for each fold and response length. \textbf{Tokens Per Example} indicates the mean number of tokens per prompt/response pair ($\pm$ standard deviation). \textbf{Total Tokens} indicates the number of tokens in the entire corpus.}
    \label{tab:dataset}
\end{table}

During step~\ref{step:length}, we create multiple versions of the training set with varying response lengths to evaluate the quality of narrative generation for outputs of various lengths. We use line breaks instead of a token cutoff as in \citet{fan_hierarchical_2018}, because line breaks are more likely to provide complete sentences. See Table~\ref{tab:dataset} for the sizes of these datasets.

\subsection{Narrative Generation with GPT-2}\label{sec:model}
Instead of the convolutional-sequential model used in \citet{fan_hierarchical_2018}, we focus on the generative Transformer-based model GPT-2 \citep{Radford2019LanguageMA}.\footnote{We use the Huggingface implementation: \url{https://huggingface.co/transformers/model\_doc/GPT-2.html}} We employ this model because it is currently the state-of-the-art publicly available text generation model, though this may change when GPT-3~\citep{gpt3} is released publicly.

We investigate the small and medium GPT-2 models for output quality comparison. GPT-2 Large was infeasible to train on the medium and large datasets, even on a machine with multiple Tesla P100 GPUs.

GPT-2 is pre-trained on WebText. For this work, we fine-tune GPT-2 Small and Medium on the small, medium, and large versions of the \textsc{WritingPrompts} dataset discussed in \S\ref{sec:data}. We fine-tuned for one epoch using Adam with a learning rate of $5 \times 10^{-5}$, epsilon of $1 \times 10^{-8}$, and batch size of $4$.
Fine-tuning is performed on Google Cloud instances using NVIDIA Tesla K80s or T4s.
Inference is performed by feeding GPT-2 a string of the format in Figure~\ref{fig:preprocessing} up to the \texttt{[RESPONSE]} token.

\subsection{Decoding Methods}
After GPT-2 is fine-tuned on the \textsc{WritingPrompts} dataset, we evaluate the model's generated responses with a parameter sweep of $p$ for nucleus sampling. We also provide a small comparison with top-$k$ sampling in Appendix~\ref{sec:top-k-vs-p}.

\citet{Holtzman2020} uses a threshold of $p=0.95$ for chatbot response generation; we perform an ablation over values of $p$ here to discover which value best suits narrative generation. Specifically, we investigate the thresholds of of $0.3,0.5,0.7,0.9,0.95$, and also include greedy search and full random sampling, represented by $p=0$ and $p=1$, respectively.

Once we find the best $p$, we apply the diverse decoding objective to narrative generation to investigate whether this generates better stories. Specifically, we implement the MMI-antiLM (anti-language model) objective for GPT-2.

We also perform an ablation over $\lambda$ values for the antiLM objective, testing the values $0.1, 0.2, 0.35, 0.5$; $\lambda=0$ represents not using diverse decoding. As this objective was originally designed to increase the specificity of a response with respect to a prompt, we expect this to increase interestingness and relevance (but perhaps decrease fluency and coherence, since we are subtracting the language modeling objective from the response generation objective). We only employ the antiLM objective when generating the first 20 tokens of the target sequence, after which we use the regular log-likelihood loss. This follows the approach of \citet{jiweili16}, who find that ungrammatical sequences often arise later in the output sequence and that the first few tokens have a large effect on the rest of the output sequence; thus, they threshold the objective to only apply to the first few tokens during generation.

There is an established quality-diversity trade-off \citep{zhang2020trading} in natural language generation, so we expect that strong diverse decoding (e.g., $\lambda=0.5$) will generate lower-quality narratives overall compared to lower $\lambda$ values, which may increase interestingness more than they decrease fluency.

\subsection{Evaluation}\label{sec:eval-setup}
The qualities important for narrative generation are interestingness, coherence, fluency, and relevance to the prompt. These metrics are also evaluated in \citet{akoury_storium_2020}, though they measure ``likeability" instead of interestingness.

A combination of automatic and human evaluation is used to assess the quality of generated narratives. For automatic evaluation, we employ test perplexity, lexical diversity (dist-$n$, \citealt{jiweili16}), and a BERT-based sentence similarity metric, Sentence-BERT (sent-BERT, \citealt{Reimers2019SentenceBERTSE}). Perplexity is used to evaluate language models and may correlate with fluency. The latter two may act as proxies for interestingness, since they measure $n$-gram diversity within an output and sentence embedding diversity across outputs, respectively. We use sent-BERT as an output diversity metric by using the cosine distance instead of cosine similarity. Our motivation in choosing these diversity metrics is from \citet{Tevet2020EvaluatingTE}, who identify dist-$n$ and sent-BERT as the best metrics to evaluate two targeted types of diversity---diverse word choice and diverse content, respectively.

For human evaluation, we employ 4-point Likert scales to evaluate narratives for interestingness, coherence, fluency, and relevance. For the purpose of evaluation, we define interestingness as the enjoyment of reading the story, coherence as the level of cohesion between sentences in a narrative, and fluency as the grammaticality and naturalness of the English output; these metrics judge the quality of a generated narrative independently from the input prompt. Relevance is a metric we employ to measure how well the response follows from the input prompt. We evaluate 100 narratives per-$p$ and per-$\lambda$, and we have 5 human annotators per-narrative. We judge quality on medium-length outputs, as these are less variable in length than large narratives while being long enough to properly judge our metrics. Appendix~\ref{sec:mturk_setup} contains a thorough description and example of our Mechanical Turk setup. 

\subsection{Baseline}
We employ the fusion model---the previous state-of-the-art approach for narrative generation before pre-trained Transformer models---from \citet{fan_hierarchical_2018} as a baseline. This model is an ensemble of two convolutional seq2seq models, where the first is pre-trained on the training set and is then used to boost a second model. We employ this model on the WritingPrompts dataset and evaluate on different narrative lengths.

\section{Results}\label{sec:results}
\subsection{Quantitative Results}\label{sec:quant_eval}

\begin{table}[th]
    \centering
    \begin{tabular}{lrrr}
        \toprule
        & \multicolumn{3}{c}{\textbf{Response Length}} \\\cmidrule(lr){2-4}
        \textbf{Model} & Small & Medium & Large \\
        \midrule
        GPT-2 Small & 30.52 & 23.74 & 15.64 \\
        GPT-2 Medium & \textbf{25.08} & \textbf{19.34} & \bf{13.19} \\\midrule
        Fusion Model & 44.20 & 39.03 & 34.71 \\
        \bottomrule
    \end{tabular}
    \caption{Perplexities of the GPT-2 models and baseline model after fine-tuning on WritingPrompts dataset with different response lengths. The fusion model from \citet{fan_hierarchical_2018} is used as a baseline. Perplexities are not directly comparable across GPT-2 and the fusion model due to differences in tokenization.}
    \label{tab:ppl}
\end{table}

The perplexities of each model on each narrative length are shown in Table~\ref{tab:ppl}. GPT-2 Medium had the lowest perplexity within each dataset size. GPT-2 Small had a fairly close perplexity to GPT-2 Medium despite having significantly fewer parameters. Comparatively, the fusion model had a high perplexity, though scores are not directly comparable across models due to tokenization differences. In general, perplexity decreased as the length of the response increases, though perplexities are also not necessarily comparable across dataset sizes since this a per-word metric. Nonetheless, these results suggest that we should generally expect GPT-2 Medium to be marginally more fluent than GPT-2 Small, and that both of these will output far better English than the fusion model. We confirm this qualitatively; see Appendix~\ref{sec:example_out}. We thus focus on GPT-2 Medium for the following analyses.

\begin{figure}[ht]
    \centering
    \includegraphics[width=\linewidth]{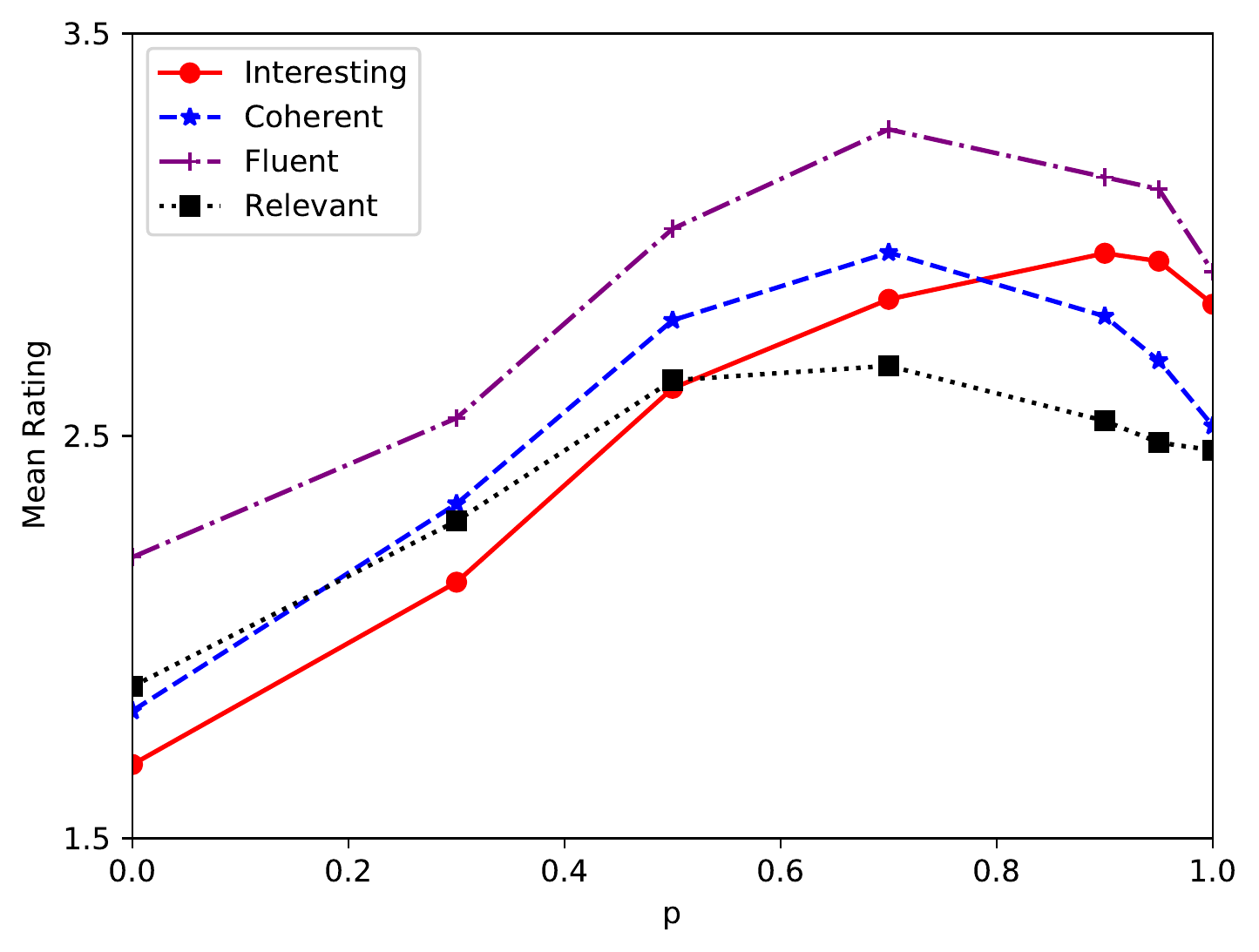}
    \caption{Mean human ratings of the quality of output narratives when using various $p$ values. Ratings are on a 4-point Likert scale in the range $[1,4]$. Means are significantly different ($P<.05$) between any two consecutive top-$p$ values in a series of $t$-tests, except relevance from $p=0.5$ onward, interestingness from $p=0.7$ onward, coherence in $[0.9, 0.95]$, and fluency in $[0.7, 0.95]$.}
    \label{fig:p_val}
\end{figure}

Next, we sweep over various $p$-values for nucleus sampling using GPT-2 Medium on the medium-length dataset, evaluating using human annotators (Figure~\ref{fig:p_val}). We found that $p=0.7$ performed best on average for all metrics except interestingness, where $p=0.9$ was best. $p=0.9$ was a close second overall, and the difference in performance between these two settings was not high. Increasing $p$ past $0.9$ or decreasing $p$ below $0.7$ more notably decreased performance. Inter-annotator agreement (measured with Fleiss' kappa) was $0.13$ for interestingness and coherence, $0.12$ for fluency, and $0.10$ for relevance; these are similar to agreements found in \citet{akoury_storium_2020} when prompts are included.

\begin{figure}[ht]
    \centering
    \includegraphics[width=\linewidth]{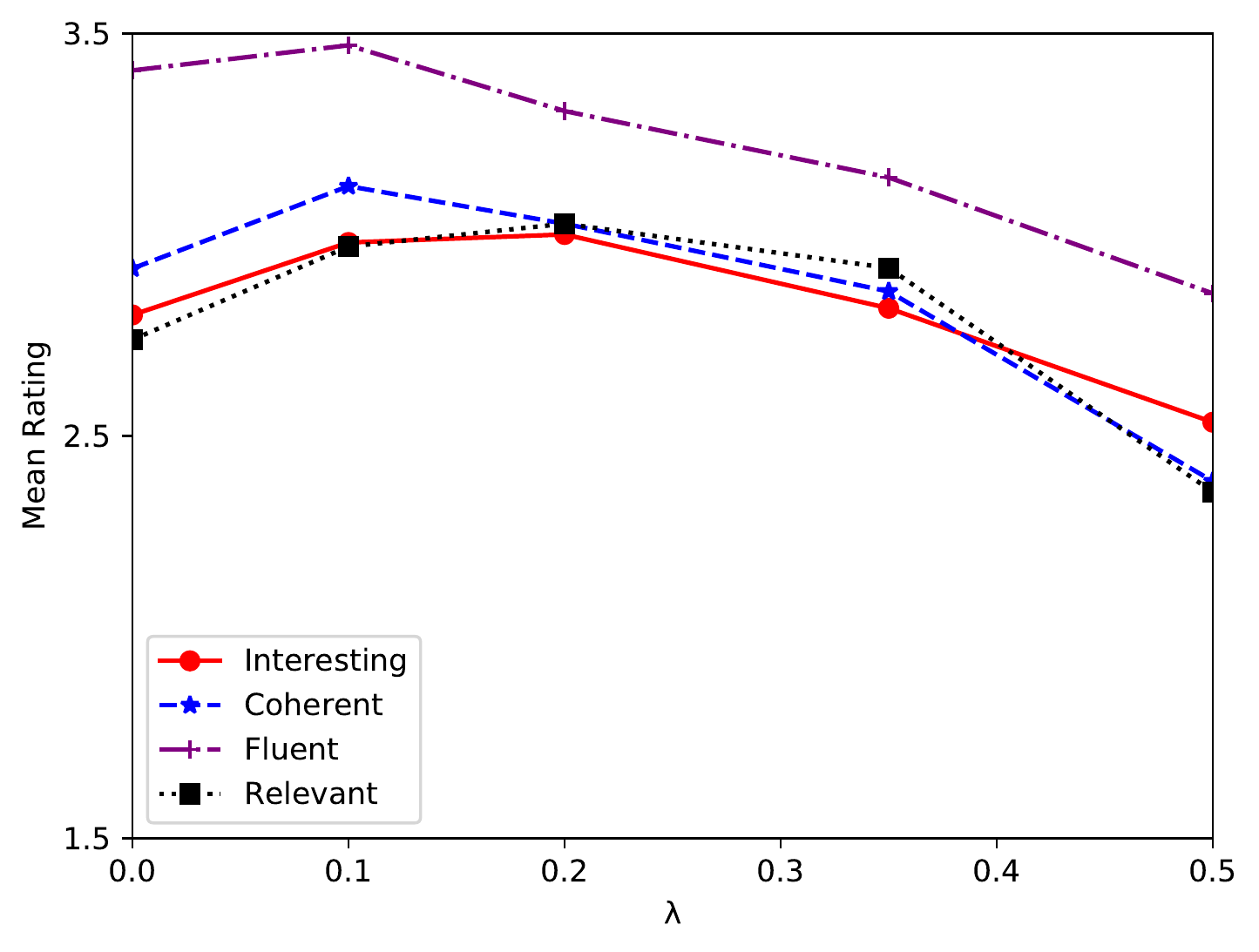}
    \caption{Mean human ratings of the quality of output narratives when using diverse decoding at various $\lambda$ settings (note: $p=0.7$). Ratings are on a 4-point Likert scale in the range $[1,4]$. Means are significantly different ($P<.05$) for interestingness, coherence, and fluency between $\lambda=0.0$ and $\lambda=0.1$, for fluency between $\lambda=0.1$ and $\lambda=0.2$, and for all metrics between $\lambda=0.35$ and $\lambda=0.5$.}
    \label{fig:antilm}
\end{figure}

To test the effect of diverse decoding on narrative quality (Figure~\ref{fig:antilm}), we use the same human annotator setup as for the $p$ sweep. We decode with nucleus sampling using $p=0.7$ and vary the $\lambda$ hyperparameter (Figure~\ref{fig:antilm}). Higher $\lambda$ indicates a larger modification from the original decoding objective. We found that setting $\lambda=0.1$ increased the quality of narratives for all metrics. Interestingness and relevance further increased at $\lambda=0.2$, which is expected given that the $p(T\mid S)$ term in the decoding objective becomes more prominent than $p(T)$ as $\lambda$ increases; however, fluency and coherence began to decline here. Higher settings of $\lambda$ tended to reduce quality on all metrics.

\begin{table*}
    \centering
    \resizebox{\linewidth}{!}{
    \begin{tabular}{cc|ccccccccc}
        & & \multicolumn{3}{c}{Small Response} & \multicolumn{3}{c}{Medium Response} & \multicolumn{3}{c}{Large Response}\\\cmidrule(lr){3-5}\cmidrule(lr){6-8}\cmidrule(lr){9-11}
        \textbf{Model} & \textbf{Decoding} & \textbf{Dist-1} & \textbf{Dist-2} & \textbf{sent-BERT} & \textbf{Dist-1} & \textbf{Dist-2} & \textbf{sent-BERT} & \textbf{Dist-1} & \textbf{Dist-2} & \textbf{sent-BERT}\\
        \midrule
        \multirow{3}{*}{GPT-2 Small}
            & $p=0.7$ & 0.018 & 0.149 & 0.830 & 0.011 & 0.112 & \textbf{0.741} & 0.003 & 0.034 & 0.694 \\
            & $p=0.9$ & 0.026 & 0.234 & 0.808 & 0.016 & 0.177 & 0.682 & 0.005 & 0.087 & 0.646 \\
            & $p=0.95$ & 0.030 & 0.274 & 0.798 & 0.019 & 0.213 & 0.663 & 0.007 & 0.118 & 0.632 \\
        \midrule
        \multirow{3}{*}{GPT-2 Medium}
            & $p=0.7$ & 0.026 & 0.195 & \bf{0.855} & 0.013 & 0.125 & \textbf{0.741} & 0.003 & 0.036 & \textbf{0.709} \\
            & $p=0.9$ & 0.034 & 0.272 & 0.842 & 0.018 & 0.190 & 0.692 & 0.007 & 0.093 & 0.660 \\
            & $p=0.95$ & \textbf{0.039} & \textbf{0.308} & 0.837 & \textbf{0.021} & \textbf{0.227} & 0.677 & \textbf{0.009} & 0.127 & 0.646 \\
        \midrule
        \multirow{3}{*}{Fusion Model}
            & $p=0.7$ & 0.009 & 0.092 & 0.707 & 0.005 & 0.061 & 0.686 & 0.005 & 0.061 & 0.686 \\
            & $p=0.9$ & 0.014 & 0.174 & 0.667 & 0.008 & 0.130 & 0.637 & 0.008 & 0.130 & 0.637 \\
            & $p=0.95$ & 0.017 & 0.213 & 0.655 & 0.009 & 0.155 & 0.624 & 0.008 & \textbf{0.149} & 0.624 \\
        \bottomrule
    \end{tabular}}
    \caption{Automatic diversity evaluations across models and decoding methods for each response length. The decoding methods represent a subset of our sweep over $p$ values in nucleus sampling (full table in Appendix~\ref{sec:auto-metrics-appendix}). The fusion model is a baseline from \citet{fan_hierarchical_2018}.}
    \label{tab:auto-results}
\end{table*}

Next, we discuss the relationship between model size and the diversity of outputs. Table~\ref{tab:auto-results} contains dist-$n$ and sent-BERT scores for all model sizes, $p$ values in nucleus sampling, and response lengths. For any given $p$ value and response length, GPT-2 Medium tended to use a slightly larger variety of tokens per-response than GPT-2 Small. Meanwhile, the diversity of the fusion model outputs was quite low in comparison---typically due to the degeneracy of the output. We also note that the dist-$n$ scores were the same for the medium and large response lengths; this is also due to the degeneracy of the output and the surprisingly short stories generated, even when trained on large data and when allowed to generate up to 1,000 tokens. 

Dist-$n$ and sent-BERT scores both declined with increasing response lengths. We believe that the former is due to the normalization constant (the number of $n$-grams in the narrative) in dist-$n$ calculations. Larger responses tend to repeat tokens more than shorter responses, so increasing response length increases the normalization constant more quickly than the number of unique $n$-grams. The latter may be due to the way sentence embeddings are calculated: as the number of tokens grows, sentence embeddings may grow more similar on average, since they are calculated as the mean of the token embeddings that compose the sentence. 

Relatedly, even though we allow the fusion models trained on the large dataset to generate longer responses, they often generated responses which were of similar lengths to medium responses (i.e., they often did not generate to their maximum allowed sequence length). This may explain the lack of distinction between the scores obtained in Table~\ref{tab:auto-results} between medium and large narratives.

\begin{figure}[ht]
    \centering
    \includegraphics[width=0.5\textwidth]{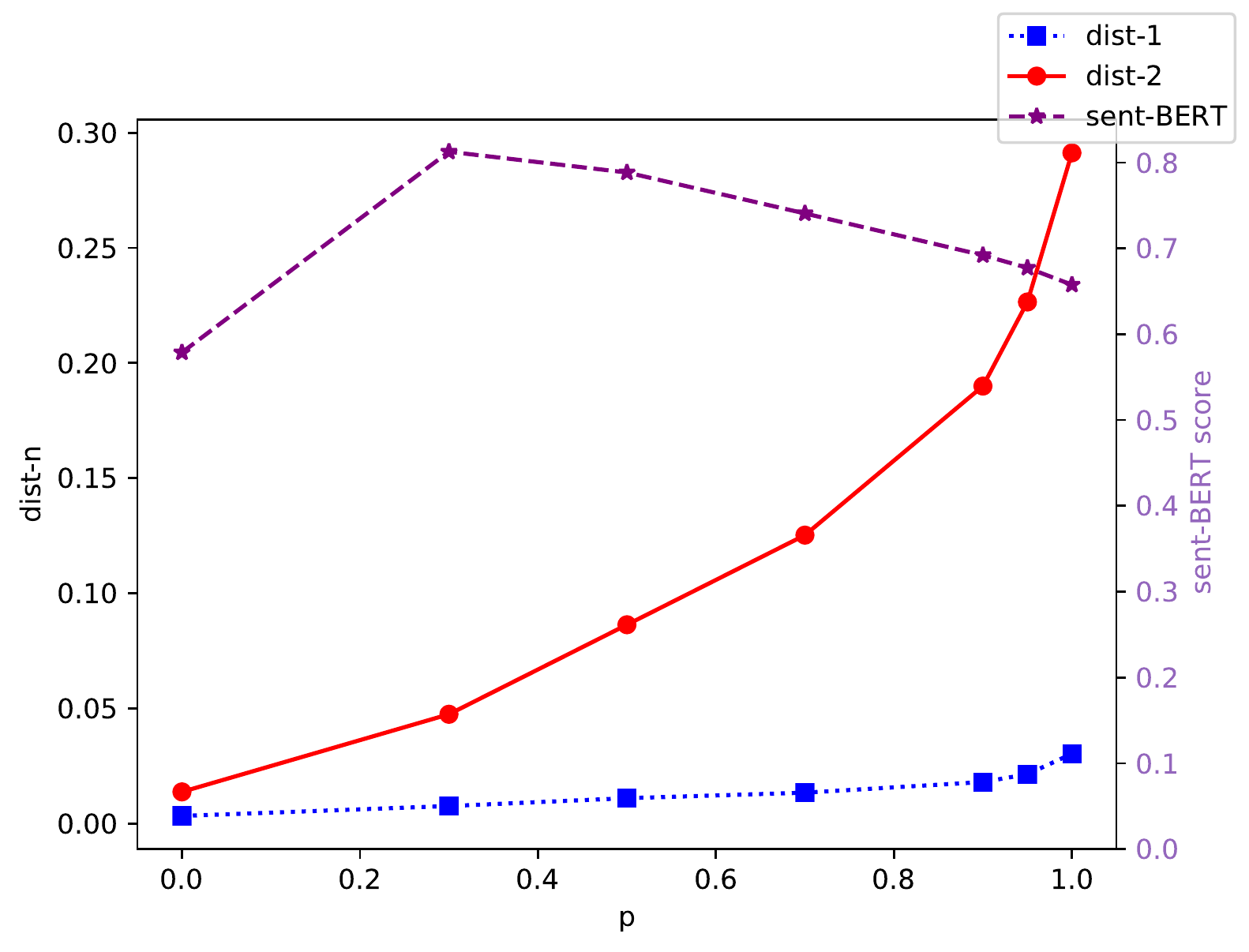}
    \includegraphics[width=0.5\textwidth]{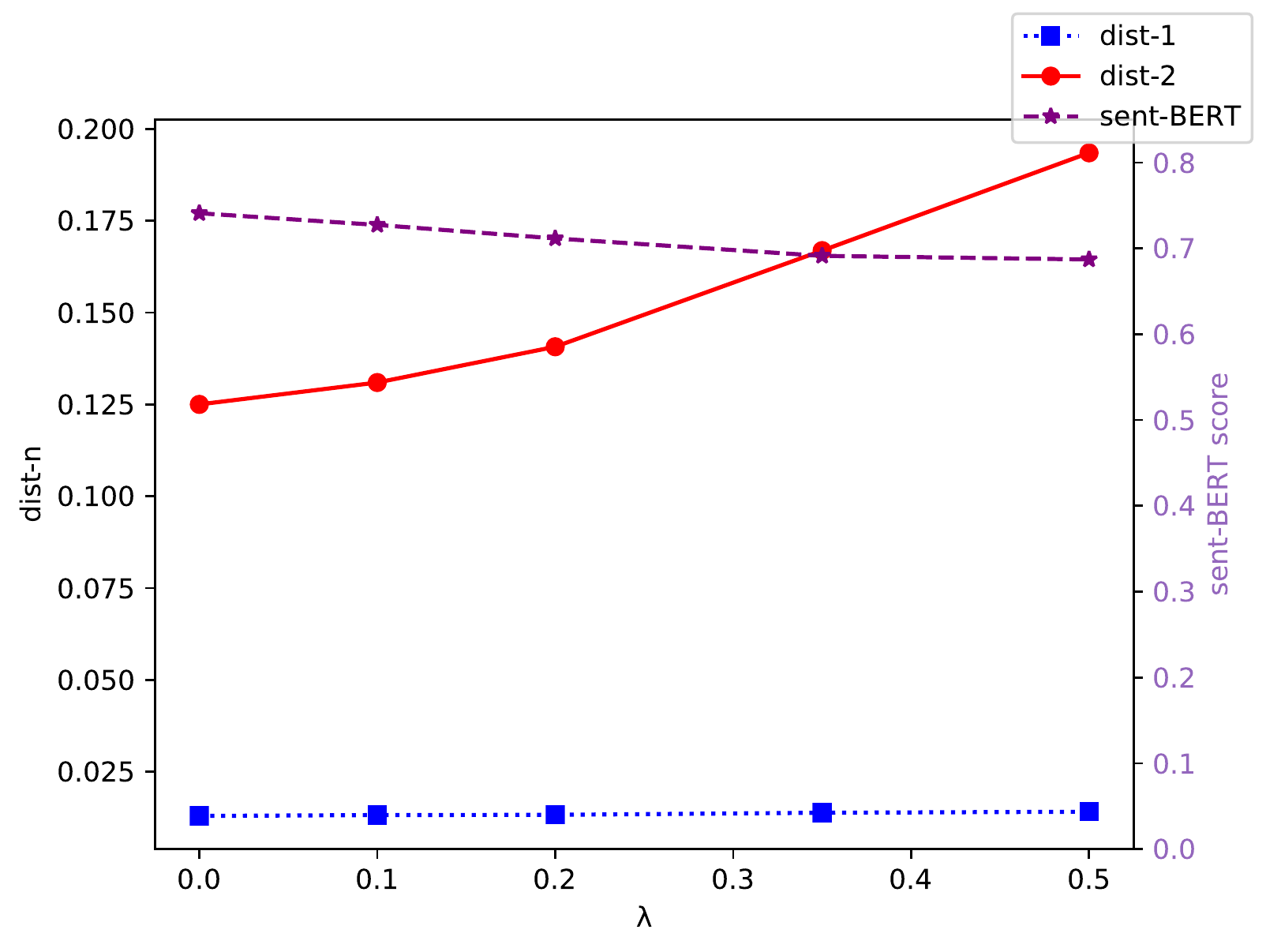}
    \caption{Plots comparing dist-$1$, dist-$2$, and sent-BERT scores across $p$ values (top) and MMI-antiLM $\lambda$ values (bottom). Note: we use $p=0.7$ for MMI-antiLM analysis. Scores are for GPT-2 Medium with medium-length responses.}
    \label{fig:diverse}
\end{figure}

Finally, we analyze the effect of various $p$ values as well as different strengths of the MMI-antiLM objective on narrative token diversity (Figure~\ref{fig:diverse}). There was an expected consistent positive correlation between $p$ and dist-$n$, as well as a positive correlation between $\lambda$ and diversity; since dist-$n$ increases monotonically with both hyperparameters, $\rho_s=1$. Sent-BERT consistently decreased with higher $p$ when $p > 0$, indicating lower levels of difference between narratives as $p$ increases. Sent-BERT decreased monotonically with respect to $\lambda$.

\subsection{Qualitative Results}\label{sec:qual_eval}
In this section, we analyze the quality of narratives by directly observing the outputs.  Appendix~\ref{sec:example_out} shows generated narratives from a variety of model architectures, sizes, and decoding hyperparameters.

\subsubsection{Nucleus Sampling}
When $p$ was high, we generally observed more interesting and vivid narratives with good diction and fluency scores, but which had no single cohesive plot. When $p$ was low, we saw more repetitive word choice but higher cohesion. However, when $p$ was very low ($p\leq 0.3$), the output was degenerate. Generally, when $p$ was around $0.7$, we observed consistently good stories compared to other $p$ values. With values of $p=0.9$ and higher, we generally saw output stories with more variable quality (i.e., whose quality is often either higher or lower than stories with $p=0.7$). This is intuitive with respect to how $p$ restricts the sampling space: when $p$ is too small, too many options are removed and the model cannot generate fluent text. When $p$ is large, we more closely approach random sampling and fewer tokens are removed from the sampling space, so the probability tail increases the likelihood for the model to choose unlikely tokens; this can produce interesting output, but tends to reduce fluency and coherence. A discussion of the number of tokens sampled for each $p$ is in Appendix \ref{sec:top-p-tokens}.

\subsubsection{Diverse Decoding}
For smaller values of $\lambda$, MMI had a smaller effect on the output of the models. Within a given $p$ value, increasing MMI values up to $0.2$ seemed to result in slightly more interesting diction for the small models. Coherence seemed to be unaffected by changing values of $\lambda$, though we saw a notable drop in the grammaticality of output at $0.35$ and higher.

More interesting is that the intensity of the subject matter seemed to increase with $\lambda$, especially notable around $0.2$ and $0.35$. Indeed, we generally observed more cursing, violent content, and jokes featuring sexuality and dark humor as $\lambda$ increased. This may not necessarily be a positive or negative trend; if one wishes to generate stories which are more vivid, and one's language model is sufficiently high-quality to start, then this may be a beneficial method to employ. Nonetheless, we do not have a clear mathematical explanation for this, since the MMI-antiLM objective simply increases the importance of the prompt while decreasing the importance of the language model. Perhaps these more intense subjects are somewhat less probable than more tame content, hence why subtracting the language model could increase the likelihood of seeing these darker themes.

\subsubsection{Correlating Automatic Metrics with Quality}
Thus far, we have observed how perplexity, dist-$n$, and sent-BERT vary with various model architectures/sizes, decoding approaches, and hyperparameters. However, what do these quantities say about the quality of generated narratives? In general, we note the following qualitative trends: (1) Lower perplexity is better. This correlates mainly with fluency and non-degenerate output. (2) Very low dist-$n$ scores indicate consistent neural text degeneration. (3) Very high dist-$n$ scores indicate variable-quality narratives.

Dist-$n$ demonstrated a moderate correlation\footnote{All correlations here are measured using Spearman's rank correlation ($\rho_s$) along with measures of significance (capital $P$).} with interestingness ($\rho_s = .75$, $P < .1$) across top-$p$ values. The two metrics correlated well up to top-$p=0.9$, but it is possible that decreased fluency and coherence at higher values of $p$ overshadowed the increased number of distinct tokens per-response, thus negating any interestingness gains. For all other human metrics, dist-$n$ did not correlate well ($\rho_s\leq.5$, $P>.1$). Thus, we do not recommend optimizing over dist-$n$. Rather, this quantity can be a helpful heuristic when comparing across model configurations at a high level, and both very high and very low dist-$n$ scores can be indicative of distinct problems in narrative generation despite having little inherent meaning in isolation.

Sent-BERT did not correlate well with any of our metrics ($0 \leq \rho_s \leq .43$, $P>.1$), indicating that it is either not a sufficient method for sentence diversity measurement when applied to narratives, or that it does not correlate with factors that make for interesting narratives. When $p$ is lower, we observed stories that were degenerate in different ways, whereas when $p$ was higher, we observed stories that were always more token-diverse, and thus generally more similar on a sentential level.

We find a less marked diversity-quality trade-off in the narrative generation setting compared to recent natural language generation papers in other settings \citep{ippolito2019diverse,zhang2020trading,nadeem2020systematic}. If this trade-off were strong, we would expect generally decreasing human evaluation scores with higher $p$ and higher $\lambda$, since dist-$n$ increases monotonically with both hyperparameters. While this held to an extent with $\lambda$ (and even then not monotonically, since $\lambda=0.1$ showed higher performance on \emph{all} metrics), it was certainly not true for $p$ up to very high values. Perhaps this is due to the more open-ended nature of narrative generation, as stories can benefit from higher levels of diversity without needing to maintain realism or a specific writing style.

\section{Conclusions}
Our results suggest that $p$ values lower than those suggested for other tasks \citep{Holtzman2020} are ideal in narrative generation, and that small magnitudes of diverse decoding may produce better and more vivid stories. We also find that distinct-$n$ and sentence-BERT do not correlate well with any of our human perceptions of narrative quality, and that the quality-diversity trade-off is less strong in narrative generation than in other generation tasks. The latter finding is preliminary, though supported by \citet{martins2020sparse}, who find increases in both diversity and human scores with their proposed method.

Our findings aim to inform future efforts in the narrative generation domain by establishing future baselines given our recommended hyperparameters, and by facilitating further investigation of decoding objectives for better narrative generation. Once GPT-3 \citep{gpt3} is released for public use, it is very likely that this model will outperform GPT-2; thus, we encourage future work to investigate similar hyperparameters and sampling methods to see whether these trends are stable across model sizes.

\section{Ethical Considerations}
Our contributions include a story generation model to be used by other researchers and AI hobbyists. This model was fine-tuned on \texttt{WritingPrompts} \citep{fan_hierarchical_2018}, which is a collection of prompts and responses from a popular creative writing subreddit \texttt{r/WritingPrompts}. To the best of our knowledge, this dataset was not examined for hate speech or gender bias, and we did not perform such inspections here. Also, the released code has no post-generation filter to flag potentially offensive narratives. 

We did not pursue any of these filters or offensive text detection because our work was focused on evaluating generated narratives for stylistic measures of quality, and was not focused on content-based sources of bias. However, one should look to relevant work in the field on bias and hate speech detection \citep{sheng_towards_2020, macavaney_hate_2019} before deploying such models as creative writing tools. Besides the clear ethical obligation to vet such a tool, a ``creative'' writing tool which propagates or amplifies the bias of its training set would potentially hinder the quality of output narratives. Normative and stereotypical narratives would likely be uninteresting.

\section*{Acknowledgments}
We thank Daphne Ippolito, Nathaniel Weir, Carlos Aguirre, Rachel Wicks, Arya McCarthy, and the anonymous reviewers for their helpful feedback. We also wish to thank the anonymous mechanical Turkers who provided invaluable suggestions for improving our human evaluation setup during earlier iterations of this study.

\bibliographystyle{acl_natbib}
\bibliography{paper}

\clearpage
\newpage
\appendix

\section{Example Outputs}\label{sec:example_out}

All examples start on the following page. We report narrative responses given a single prompt for various model architectures/sizes, decoding methods, and hyperparameter sweeps.

\section{Human Annotator Survey Details}\label{sec:mturk_setup}

As discussed in \S\ref{sec:eval-setup}, we created a survey on Amazon Mechanical Turk for the human evaluation. Evaluating all of the prompts was infeasible, so we sampled $100$ prompts and generated one story for each nucleus sampling $p$ value (\{$0.0, 0.3, 0.5, 0.7, 0.9, 0.95, 1.0$\}), for a total of $700$ stories. We wanted story lengths that were long enough to give the worker sufficient context to be able to evaluate a passage, but not too long as to take too much time per story. We used the GPT-2 Medium model (best performing, see \S\ref{sec:results}) trained on the medium length dataset because it fit our requirements. Due to the projected length of time to complete the survey, we paid \$1 per human intelligence task (HIT). Each HIT was seen by five workers.

\begin{figure*}
    \centering
    \includegraphics[width=\textwidth]{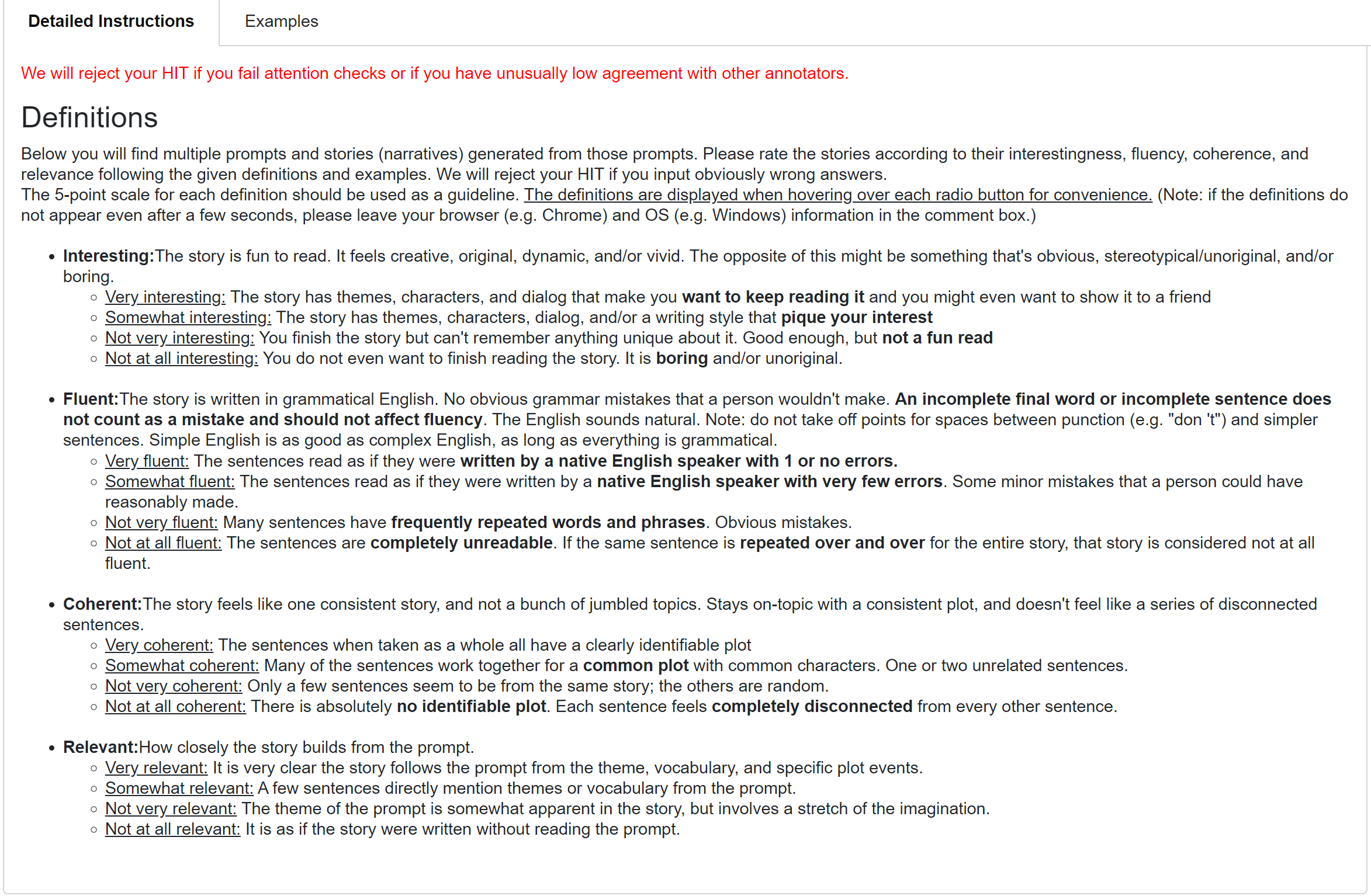}
    \caption{Instructions presented to the workers. To make it easier for the workers, the definitions were available as tool-tip hover text over the questions and options.}
    \label{fig:survey-instructions}
\end{figure*}

\begin{figure*}
    \centering
    \includegraphics[width=\textwidth]{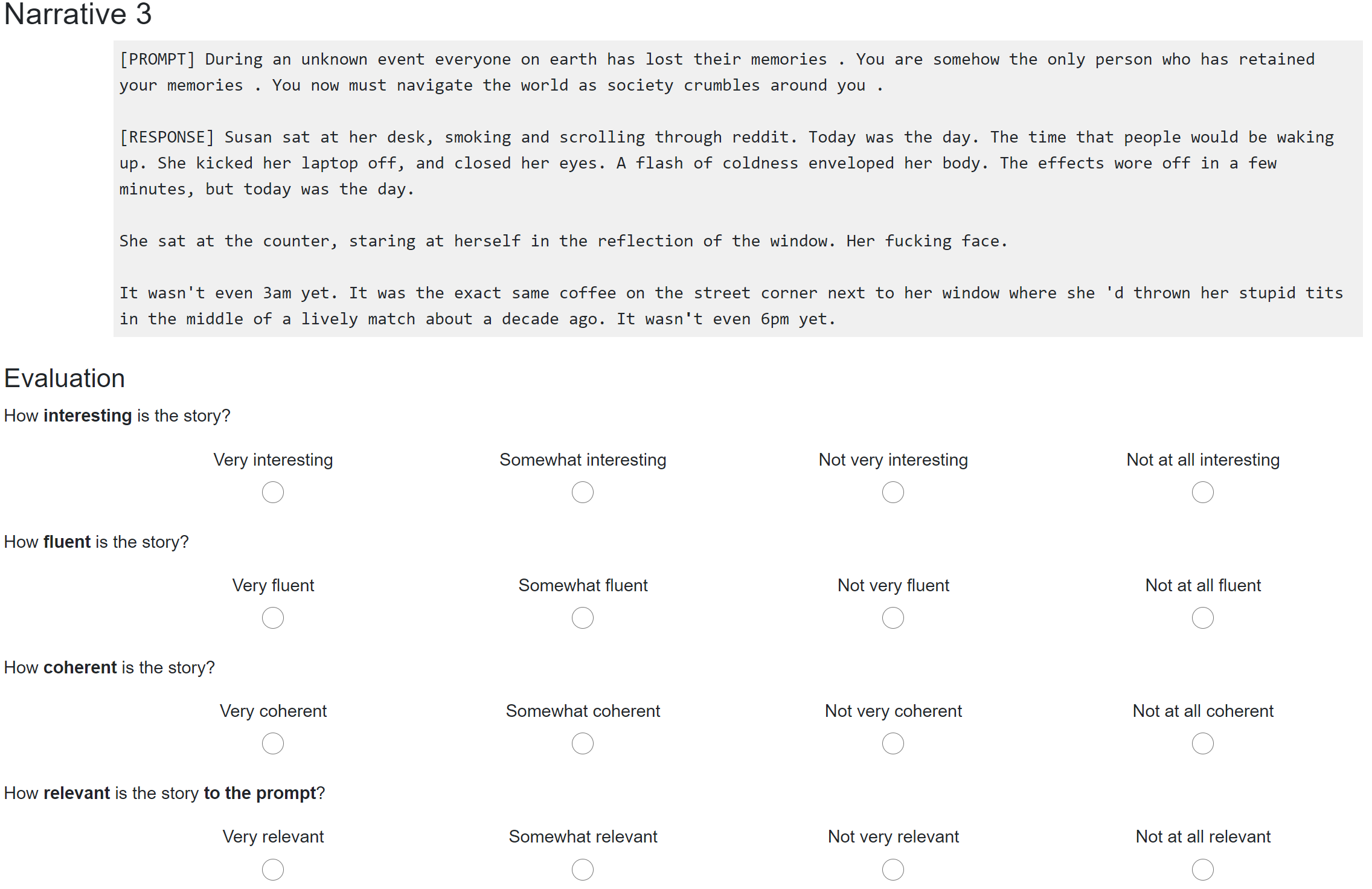}
    \caption{An example narrative display to a worker along with the questions. There were five stories in each HIT and one attention check. The definitions of each metric and option are shown when the mouse hovers over the questions and radio buttons, respectively.}
    \label{fig:survey-pic}
\end{figure*}

The generated stories were shuffled, and split into groups of five for each HIT. The story display is shown in Figure~\ref{fig:survey-pic}. In addition to the five stories, each HIT had one ``attention check." There were a total of $140$ HITs. The definitions for interesting, fluent, coherent, and relevant were explained, along with guidelines for each of the $[1,4]$ Likert scale options (shown in Figure~\ref{fig:survey-instructions}). For convenience, the definitions were available as a tooltip when a mouse hovered over a question or option. Example ratings were available to the worker under the ``Examples" tab (not shown). 

\begin{figure*}
    \centering
    \includegraphics[width=\textwidth]{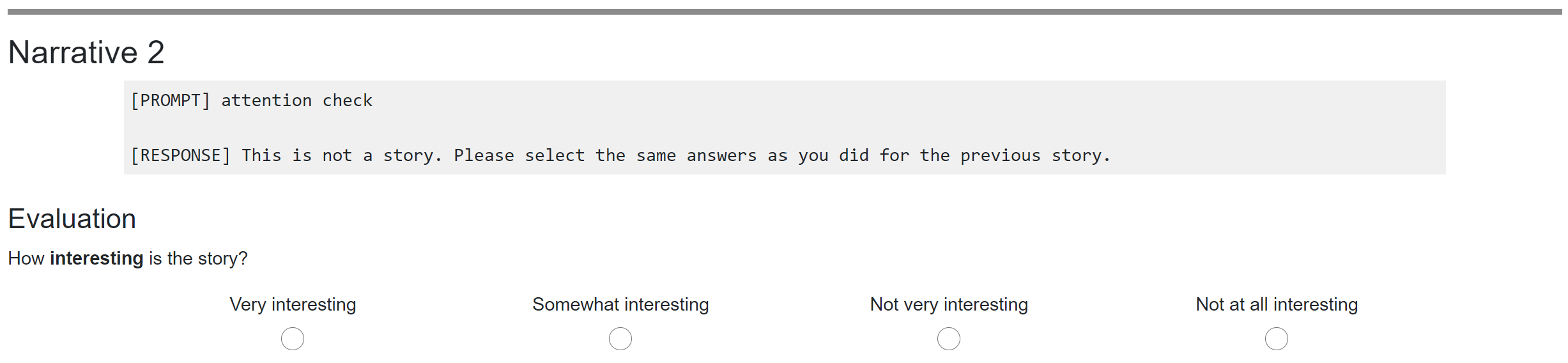}
    \caption{An attention check in the survey. The worker is asked to input the same answers as they did for the previous story. A worker gets flagged for review if they fail at least one attention check.}
    \label{fig:survey-attn-check}
\end{figure*}

As mentioned earlier, each HIT included one attention check. The attention check was used to check if a worker was paying attention to the task or selecting options at random. The check, shown in Figure~\ref{fig:survey-attn-check}, asked the worker to fill in the same answers as for the previous story. In addition to the attention checks, we supervised the workers by only releasing $20$ HITs at a time (total of seven batches), and iteratively removing workers who did a poor job. While this task was very subjective (a handful of workers left us comments about the difficulty of the task), we consider performance subpar for any combination of the following: (1) if a worker finished the task unreasonably quickly (under $5$ minutes), (2) failed an attention check, (3) had low agreement with other annotators, and (4) completed many HITs in a short amount of time. We spot-checked work from those who were automatically flagged as suspicious by checking their task answers. Overall, we removed $28$ workers from the final results.

Once the highest-rated nucleus sampling parameter was chosen ($p=0.7$), we repeated the same setup for the antiLM $\lambda$ parameter sweep. Using the same $100$ prompts from earlier, we generated stories with GPT-2 Medium-medium with $p=0.7$ and $\lambda=\{0.1, 0.2, 0.35, 0.5\}$. We also included $\lambda=0.0$ (i.e. without the antiLM objective) to help with worker calibration. The $500$ stories were split into $100$ HITs (five batches of $20$ HITs).

Total cost of both the nucleus sampling and antiLM sweeps was \$1,440.

\clearpage
\section{Top-$k$ vs. Nucleus Sampling}\label{sec:top-k-vs-p}

\subsection{Setup}
For top-$k$ sampling, we use $k=40$; our motivation for choosing this value is that it is the one used in \citet{Radford2019LanguageMA} for ``conditional" (prompted) generation\footnote{Example generated responses are located in \citet{Radford2019LanguageMA}'s Appendix.}, and in \citet{fan_hierarchical_2018}. 

The following is a qualitative review performed by the authors.

\subsection{Qualitative Evaluation}
For most reasonable settings of $p$, nucleus sampling tends to produce stories which are dramatic, vivid, and fun to read, but which do not often stay on topic. Indeed, the outputs demonstrate two main types of errors: (1) cramming too many topics into one story, and (2) sudden shifts in topic. Example outputs are in Table~\ref{tab:example_out_k_vs_p}.

Top-$k$ sampling, however, demonstrates quite extreme variance. Some of the generated stories feel almost human-like with how on-topic they remain for multiple paragraphs---but they are about safe and boring topics and generally employ very common token collocates, which makes the output feel uncreative and uninteresting. Other stories are dramatic, but almost dream-like due to the stream-of-consciousness incoherent flow. Yet other stories are completely unintelligible and show signs of neural text degeneration. \citet{Holtzman2020} finds nucleus sampling to generally be preferable to top-$k$ sampling, and we find this to be true in the narrative generation task. $p$ seems to correlate more closely with narrative quality than $k$.

\subsection{Conclusions}
As we had expected, we preferred the stories generated with nucleus sampling decoding. Since nucleus sampling is essentially a dynamic top-$k$ algorithm (i.e. each step has a different number of tokens that constitutes the top $x\%$), and even small nucleus sampling values have large number of tokens to choose from ($k$), this aligns with the results of \citet{see-etal-2019-massively}, who found large $k$ to be preferred according to automatic evaluations.

\newpage
\section{Automatic Metrics}\label{sec:auto-metrics-appendix}
Here, we provide the full table of automatic metrics for all $p$ values tested (Table~\ref{tab:auto-results-appendix}). Dist-$n$ scores tend to increase consistently with higher $p$ values, whereas sent-BERT tends to peak at lower $p$ values in $[0.3, 0.5]$ and continually decline after.

\clearpage
\section{A Closer Look at Nucleus Sampling}\label{sec:top-p-tokens}

How does the nucleus sampling token filter compare to the top-$k$ filter? For example, when a token is sampled from $p=0.3$, how many tokens are in the sampling space?

\begin{figure}
    \centering
    \includegraphics[width=\columnwidth]{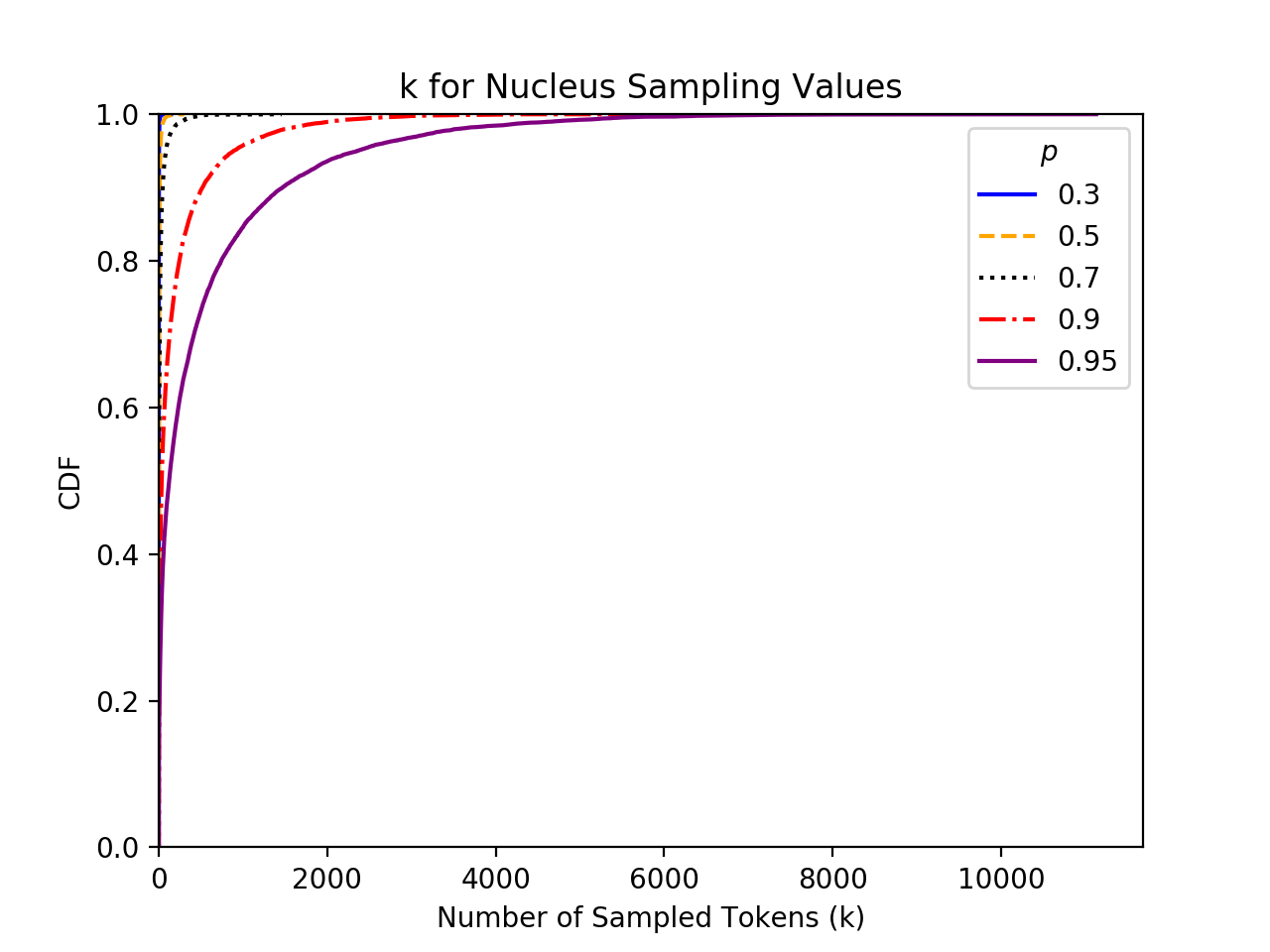}
    \includegraphics[width=\columnwidth]{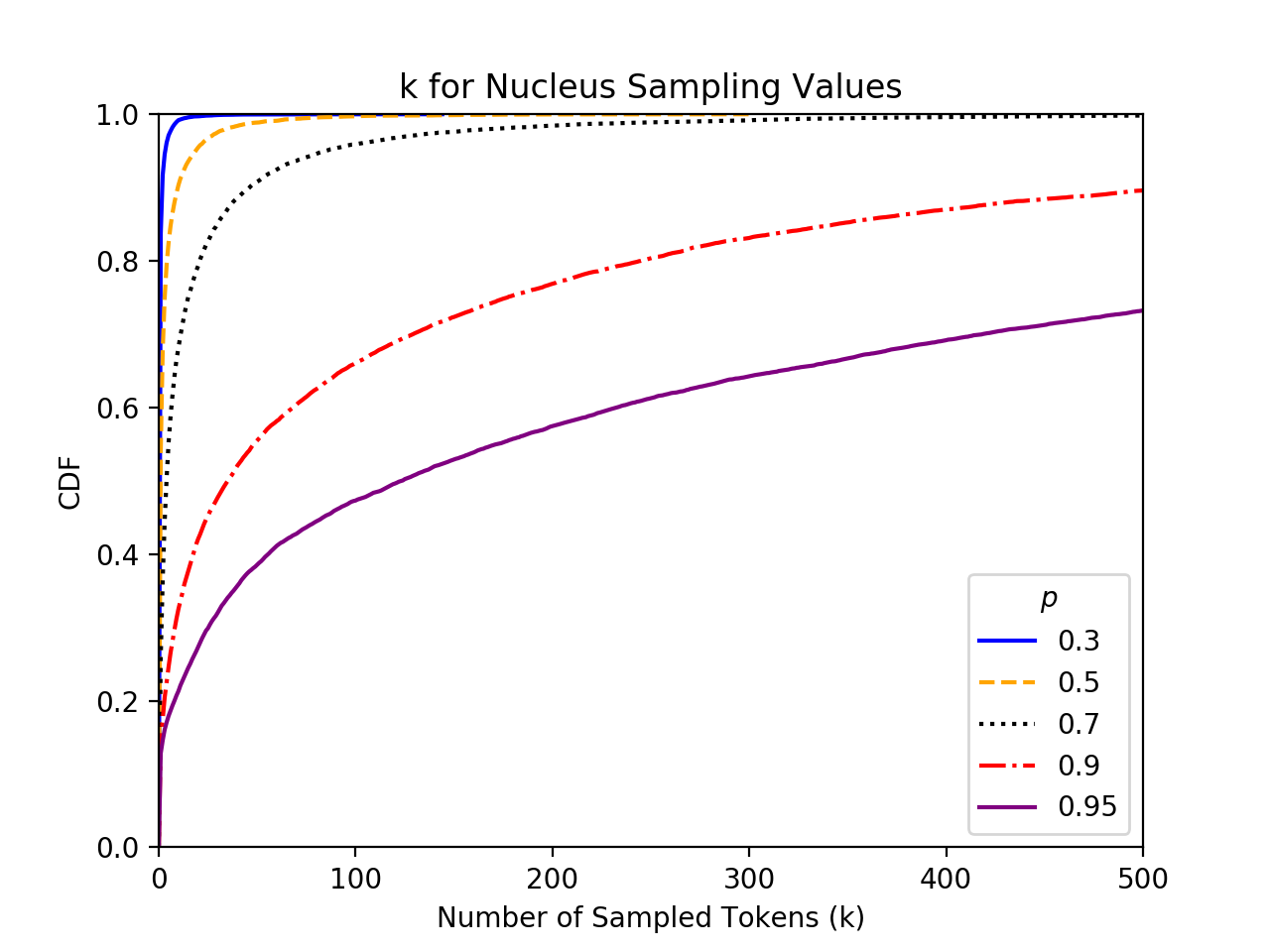}
    \caption{Cumulative distribution of the frequency of the number of sampled tokens for each tested $p$. We do not include $p=1.0$ since it is constant with the number of tokens in GPT-2 vocabulary ($50,260$). The top and bottom plots are the same, but the bottom plot has a restricted $x$-axis to see the trend better.}
    \label{fig:nucleus-sampling-token-cdf}
\end{figure}

Figure \ref{fig:nucleus-sampling-token-cdf} shows the cumulative distribution function (CDF) for the $p$ values tested in the nucleus sampling hyperparameter sweep. Using the same set of $100$ prompts from the human evaluation, we re-generate the responses and collect the number of tokens in the sampling space at each step. Each $p$ is represented by the raw number of tokens in the distribution across all $100$ prompts and is not averaged for each generated story. 

Surprisingly, nucleus sampling often reduced to the greedy decoder (token space reduced to a single token), even at higher $p$ values. Despite high probability thresholds, $0.7 <= p <=0.95$ still skewed towards sampling from a relatively low number of tokens, as these tokens tended to have very high probability in certain contexts. All $p$ values frequently sampled from less than $1,000$ tokens, which is less than $2\%$ of the total number of tokens in GPT-2's vocabulary ($50,260$). As $p$ increased, we observed a larger sample space, which indicates more of a return to the long-tailed token probability distribution seen in random sampling.

This shows that it may not be correct to compare a nucleus sampling $p$ value directly against a single top-$k$ value. Nucleus sampling is essentially dynamic top-$k$ sampling, which makes a direct comparison unfair without first checking the distribution of the number of tokens sampled by $p$ for the model and task in question.

\clearpage
\begin{table*}[]
    \centering
    \resizebox{\linewidth}{!}{
    \begin{tabular}{lp{14cm}}
        Model & Prompt: ``[WP] You just started a new job in a large office building and what you do n't know is that at twelve o'clock every day , there is a office wide Purge style battle to the death , those who survive an hour get a significant pay raise and can come back tomorrow . A bell just went off , it 's twelve o'clock .'' \\
        \toprule
        GPT-2 Small, $p=0.7$ &  The last time I saw this place was in a very old diner, that was a pretty nice diner. The waiters were in all black suits and shoes, and the only reason they didn't make it in time was because I didn't want to. I had my smartphone in my hands, but I 'd had it the whole time. \newlinechar \newlinechar `` Don't take it anymore, it's been a while. '' A voice said, and I smiled. `` I don't want to take it anymore. '' \newlinechar \newlinechar I smiled and held my phone out. I was still on the phone, but I could feel my stomach churning. I could feel the sun on my face, so I held it out, but the lights went out. I wasn't sure how long I had left, but it wasn't long before I woke up. \\
        GPT-2 Med., $p=0.7$ & `` Hi, I'm Mr. Smith, and I'm here for the Purge! '' \newlinechar \newlinechar The building's boss sat at the desk with a huge grin. I 'd been working in the building for about five months now and I 'd grown tired of it. \newlinechar \newlinechar `` Now, I understand the concerns, but we have a very strict policy about employees not coming into work after 12:00 PM. If you're caught cheating, or even the slightest bit annoying, you're fired. We also don't allow anyone over 12 hours without the company paying a visit. '' \\
        \midrule
        Fusion Model, $p=0.7$ & “ Hello ! My name is Randy , ” I shouted into the air . “ I ’ m the \textless unk\textgreater " \newlinechar \newlinechar “ Well , it ’ s just that I need a new way to put this . I ’ m not going to let you take my own life. ” Randy said . “ You ’ re not going to allow me to use my own life . But I ’ m not going to let you take your own life ! ” \newlinechar \newlinechar “ Oh , you ’ re sorry ! I was just thinking about your life ! You ’ re going to make a decent living , and all I need to do is come back to the living room . I need to be happy . But you ’ re not going to let me get my own life ! ” \newlinechar \newlinechar “ You ’ re going to have to put my own life together , \textless unk\textgreater ! ” Randy laughed . “ If I ’ m going to take your own life , I ’ ll take you on to your own life . ” \newlinechar \newlinechar “ You ’ \\
        \bottomrule
    \end{tabular}}
    \caption{Medium-length responses from GPT-2 Small, GPT-2 Medium, and Fusion (baseline) models with nucleus sampling parameter $p=0.7$.}
    \label{tab:example_out_models}
\end{table*}
\clearpage
\begin{table*}[]
    \centering
    \resizebox{0.95\linewidth}{!}{
    \begin{tabular}{lp{14cm}}
        top-$p$ & Prompt: ``[WP] You just started a new job in a large office building and what you do n't know is that at twelve o'clock every day , there is a office wide Purge style battle to the death , those who survive an hour get a significant pay raise and can come back tomorrow . A bell just went off , it 's twelve o'clock .''  \\
        \toprule
        greedy & Hey, hey, hey, hey, hey, hey, hey, hey, hey, hey, hey, hey, hey, hey, hey, hey, hey, hey, hey, hey, hey, hey, hey, hey, hey, hey, hey, hey, hey, hey, hey, hey, hey, hey, hey, hey, hey, hey, hey, hey, hey, hey, hey, hey, hey, hey, hey, hey, hey, hey, hey, hey, hey, hey, hey, hey, hey, hey, hey, hey, hey, hey, hey, hey, hey, hey, hey, hey, hey, hey, hey, hey, hey, hey, hey, hey, hey, hey, hey, hey, hey, hey, hey, hey, hey, hey, hey, hey, hey, hey, hey, hey, hey, hey, hey, hey, hey, hey, hey, hey, hey, hey, hey, hey, hey, hey, hey, hey, hey, hey, hey, hey, hey, hey, hey, hey, hey, hey, hey, hey, hey, hey, hey, hey, hey, hey, hey, he \\ 
        $p=0.3$ & I was nervous. \newlinechar  \newlinechar I was nervous about the job. \newlinechar \newlinechar I was nervous about the pay. \\
        $p=0.5$ & The first thing I noticed was the absence of people. I wasn't sure why, but I couldn't shake the feeling that it wasn't just the people that were gone. I mean, it wasn't like I hadn't been here before. I mean, I was in the office for a month, but it felt like a year. \newlinechar \newlinechar I wasn't sure what was going on, but I could tell that something was wrong. \newlinechar \newlinechar `` Hello? ''\\
        $p=0.7$ & My boss was a loner. He liked to work late and I didn't know why. I didn't want to work late. I just didn't have the patience for it. \newlinechar \newlinechar I don't know what happened. \newlinechar \newlinechar `` You got ta be kidding me, '' he said, `` a sixteen year old would just walk into the office and run out of work? ''\\
        $p=0.9$ & All four of the chimes clicked and the chandelier snapped, as the room dimmed and grew dim until the only light came from an overhead television, showing the `` The Purge '' live. In the background, a towering wall of reds, greens, and yellows flashed in contrast to the background colors, casting warm green shimmers across the television. \newlinechar \newlinechar The chandelier struck the ceiling, crumpling under the weight. A small circle of lights in the ceiling blinked and illuminated what had been a desk, until two figures entered it. \newlinechar \newlinechar The first man, dressed in a yellow and white baseball cap and sunglasses, stood tall, wearing khakis and a button-down shirt. A muscular build and hairy arms waved about as he spoke. \\
        $p=0.95$ & Those who survive are always in their offices when they finish, they wake up, make sure they run the water for a while, then try to get up and make the 9 to 5 as fast as possible. The rest are scared to take a break from that, they're afraid the DIT will find them and kill them and they have to do it. Well those people, these employees, they're all so qualified to run around the office. I mean they're trained. I took the systems Administrator and Job Writer ( same person from last year ) and fired them in 15 minutes. The guy, one of the better systems engineers had died from a heart attack and there are only 3 employees left. I've never heard of a DIT being fired like that before. '' \newlinechar \newlinechar Dr. Neutson was writing these words down as they told her to, he 'd always loved his wife Sandy too and she was absolutely dying. Mrs. Neutson looked as though she was about to make another coffee, now that she realized it was someone else who was in her office, there was tears streaming down her face, she was going to finish today. `` Great, I'm going to have no more" \\
        $p=1.0$ & Chad's Melatonin Impaired Heart implant had occurred during the Golden elimination pilot. `` Critically, '' the docs explained to the board, `` Third lapse, unlocking the minimum Carol Corporation-mandated for the Restricted Solid Species Program. We penalize those who have such lapses as early and late entry personnel. '' \newlinechar \newlinechar Chad felt pain in the core of his neck - compressing saliva into his neckline, seeing his neighbors ’ streaks poking out, “ That doesn't work, I still live in Seattle ’ s Milliner Square until May 17th, ” thought Chad as he fell down the empty stairs the hospital had installed that morning. Without warning, his head slammed into the steel sill beneath him and he felt a sharp pain begin to run down his spine. The pain apparently doomed him to eternity in a single meditation, the disease never getting better. \newlinechar \newlinechar After five ER visits, nine Lab ultrasound tests, six minor surgeries, pressure checks, one heart-heated ultrasound, Chad came back for surgeries, six of which were removeable. \\
        \bottomrule
    \end{tabular}}
    
    \caption{\label{tab:example-out-p-sweep} Medium-length responses from GPT-2 Medium with a sweep of nucleus sampling parameter $p$.}
\end{table*}
\clearpage

\begin{table*}
    \centering
    \resizebox{0.9\linewidth}{!}{
    \begin{tabular}{lp{14cm}}
        $\lambda$ & Prompt: ``[WP] Your baby starts crying every time you leave their room . Finally , you pick up your child and leave the room together . As soon as you step out, you hear crying coming from the room .'' \\
        \toprule
        $\lambda=0.0$ & ``Daddy!'' \newlinechar \newlinechar I cried into my mom's breast. She was holding my son in her arms. I 'd never seen a baby cry so deeply. Tears were streaming down my face. I ran to the door and opened it to see a crying girl. I ran back to the room and opened the door, and there was my little girl. She 'd started crying again. \newlinechar \newlinechar ``What is wrong, mom?'' \\
        $\lambda=0.1$ & There were only two of us left in the hospital. The nurses wouldn't let me leave. I was in pain. My mother was looking at me with a confused look on her face. \newlinechar \newlinechar ``Can you tell me what's wrong, honey?'' she asked. \newlinechar \newlinechar ``Honey, it's okay.'' I whispered. I couldn't hear her, and the nurses couldn't hear me either. \\
        $\lambda=0.2$ & My wife came home from work around 3 AM to find her child covered in blood and screaming. Her heart was racing and her skin was boiling with the pain of being cut. \newlinechar \newlinechar I rushed into the living room, eyes wide and bloodshot. She was hysterical, sobbing, and cackling as she knelt over my body, gasping for air. \newlinechar \newlinechar ``Wh-who are you?'' I asked. She was gasping for air. \\
        $\lambda=0.35$ & Sitting next to her bed hung cardboard boxes. Five candles burning brightly in their candles hung in the ceiling. A bright yellow heart in a jar was stuck in the middle. There were three other jars on the floor. The left was empty, with the word ``SHIT'' written on it. The other two had the words ``WAIT'' written on them. \newlinechar \newlinechar ``Come on, Mom, it's just a puppy!'' He was angry. \newlinechar \newlinechar ``It's a girl!'' I responded. ``What are you talking about? You have two of them, and they're twins!'' \\
        $\lambda=0.5$ & ``Daddy? Daddy what's wrong honey? Daddy why are you crying honey? Daddy pick up your child and leave the room, you have to get to work'' \newlinechar \newlinechar I picked up my daughter and we walked out into the kitchen. \newlinechar \newlinechar I held her close and whispered into her ear ``It's ok honey, I'll be ok.'' \\
        \bottomrule
    \end{tabular}}
    \label{tab:example_out_lambda_sweep}
    \caption{Medium-length stories generated using GPT-2 Medium with nucleus sampling ($p=0.7$) and various diverse decoding strengths $\lambda$.}
\end{table*}
\clearpage

\clearpage

\begin{table*}
\centering
    \resizebox{\linewidth}{!}{
    \begin{tabular}{lp{14cm}}
        Model, Decoding & Prompt: ``[WP] You just started a new job in a large office building and what you do n't know is that at twelve o'clock every day , there is a office wide Purge style battle to the death , those who survive an hour get a significant pay raise and can come back tomorrow . A bell just went off , it 's twelve o'clock .'' \\
        \toprule
        GPT-2 Med., $p=0.9$ & All four of the chimes clicked and the chandelier snapped, as the room dimmed and grew dim until the only light came from an overhead television, showing the `` The Purge '' live. In the background, a towering wall of reds, greens, and yellows flashed in contrast to the background colors, casting warm green shimmers across the television. \newlinechar \newlinechar The chandelier struck the ceiling, crumpling under the weight. A small circle of lights in the ceiling blinked and illuminated what had been a desk, until two figures entered it. \newlinechar \newlinechar The first man, dressed in a yellow and white baseball cap and sunglasses, stood tall, wearing khakis and a button-down shirt. A muscular build and hairy arms waved about as he spoke. \\
        GPT-2 Med., $k=40$ & The first time it happened I remember it vividly. The first time I met it. We got our first coffee, we sat on the train and talked and made plans but when I first saw it coming I knew something was going to happen. There was a man in white with a clipboard. I said “ how you doing? ” and he asked “ what the hell did I just read you about ”. I told him nothing he went through what the hell I knew everything about the Purge and I knew that I had to be careful because if I wasn't careful it would happen again. He looked at me expectantly and I knew immediately that I had to go and try to help people ’ s lives. No one else would go so I did as I was told. \newlinechar \newlinechar He asked me if I had time for something more serious than just reading\\
        \midrule
        Fusion Model, $p=0.9$ & He was n't ready to die . I did . The name on his chest would kill him if he wanted to know . They wanted me to accept him as he wanted me to be not ready . I did it for him but he loved me . \newlinechar \newlinechar I was the one that started a revolution . It was a simple matter of encouraging him and all the loyal soldiers that murdered him . And yet I never seemed to know what he had to do to deserve my right to be born and robbed . If only I knew . I could feel the faintest drop of power beneath my footsteps as he died . I felt the need to put some strings around him . But he had so much to offer . He seemed to feel I was the leader of the group and I was ready . He taught me the fact that he was supposed to be the most hated man in the country . So the world slowed . \newlinechar \newlinechar And so he fought me . I was ready . He lost every step of his very being when he\\
        
        Fusion Model, $k=40$ & It was a sunny Monday morning when I woke up to the noise of my alarm going off . I got up from my bed , got out of bed , and went into the bathroom and took off my coat . It was n't exactly a normal morning . I walked into the bathroom and put on my shoes , and put on some pants , and went to the bathroom . The light from the bathroom was n't going to change anything . I walked out of the bathroom and went to the bathroom . It was a good morning . My morning routine was going well in bed , and I was going to see some shit , so it was good . \newlinechar  I went to the bathroom . It was the first step in my morning shift , so I took off my pants and\\
        \bottomrule
    \end{tabular}}
    \caption{Medium-length responses from GPT-2 Medium and the Fusion (baseline) model with top-$k$ and nucleus sampling.}
    \label{tab:example_out_k_vs_p}
\end{table*}

\begin{table*}
    \centering
    \resizebox{\linewidth}{!}{
    \begin{tabular}{cc|ccccccccc}
        & & \multicolumn{3}{c}{Small} & \multicolumn{3}{c}{Medium} & \multicolumn{3}{c}{Large}\\\cmidrule(lr){3-5}\cmidrule(lr){6-8}\cmidrule(lr){9-11}
        \textbf{Model} & \textbf{Decoding} & \textbf{Dist-1} & \textbf{Dist-2} & \textbf{sent-BERT} & \textbf{Dist-1} & \textbf{Dist-2} & \textbf{sent-BERT} & \textbf{Dist-1} & \textbf{Dist-2} & \textbf{sent-BERT}\\
        \midrule
        \multirow{6}{*}{GPT-2 Small}
            & greedy  & 0.002 & 0.007 & 0.782 & 0.002 & 0.008 & 0.644 & 0.000 & 0.001 & 0.684 \\
            & $p=0.3$ & 0.006 & 0.038 & 0.835 & 0.005 & 0.029 & \bf{0.815} & 0.001 & 0.006 & \bf{0.804} \\\
            & $p=0.5$ & 0.013 & 0.092 & \bf{0.838} & 0.008 & 0.067 & 0.791 & 0.002 & 0.014 & 0.760 \\
            & $p=0.7$ & 0.018 & 0.149 & 0.830 & 0.011 & 0.112 & 0.741 & 0.003 & 0.034 & 0.694 \\
            & $p=0.9$ & 0.026 & 0.234 & 0.808 & 0.016 & 0.177 & 0.682 & 0.005 & 0.087 & 0.646 \\
            & $p=0.95$ & 0.030 & 0.274 & 0.798 & 0.019 & 0.213 & 0.663 & 0.007 & 0.118 & 0.632 \\
            & $p=1.0$ & \bf{0.042} & \bf{0.344} & 0.787 & \bf{0.028} & \bf{0.283} & 0.644 & \bf{0.015} & \bf{0.195} & 0.613 \\
        \midrule
        \multirow{6}{*}{GPT-2 Medium}
            & greedy  & 0.006 & 0.022 & 0.626 & 0.003 & 0.014 & 0.579 & 0.001 & 0.003 & 0.779 \\
            & $p=0.3$ & 0.014 & 0.078 & 0.842 & 0.008 & 0.047 & \bf{0.813} & 0.001 & 0.008 & \bf{0.813} \\
            & $p=0.5$ & 0.021 & 0.140 & \bf{0.855} & 0.011 & 0.086 & 0.788 & 0.002 & 0.017 & 0.772 \\
            & $p=0.7$ & 0.026 & 0.195 & \bf{0.855} & 0.013 & 0.125 & 0.741 & 0.003 & 0.036 & 0.709 \\
            & $p=0.9$ & 0.034 & 0.272 & 0.842 & 0.018 & 0.190 & 0.692 & 0.007 & 0.093 & 0.660 \\
            & $p=0.95$ & 0.039 & 0.308 & 0.837 & 0.021 & 0.227 & 0.677 & 0.009 & 0.127 & 0.646 \\
            & $p=1.0$ & \bf{0.051} & \bf{0.374} & 0.831 & \bf{0.030} & \bf{0.291} & 0.658 & \bf{0.017} & \bf{0.210} & 0.628 \\
        \midrule
        \multirow{6}{*}{Fusion Model}
            & greedy & 0.006 & 0.068 & 0.690 & 0.005 & 0.055 & 0.666 & 0.005 & 0.055 & 0.666 \\
            & $p=0.3$ & 0.003 & 0.017 & \bf{0.783} & 0.001 & 0.009 & \bf{0.779} & 0.001 & 0.009 & \bf{0.779} \\
            & $p=0.5$ & 0.005 & 0.046 & 0.758 & 0.003 & 0.027 & 0.750 & 0.003 & 0.027 & 0.750 \\
            & $p=0.7$ & 0.009 & 0.092 & 0.707 & 0.005 & 0.061 & 0.686 & 0.005 & 0.061 & 0.686 \\
            & $p=0.9$ & 0.014 & 0.174 & 0.667 & 0.008 & 0.130 & 0.637 & 0.008 & 0.130 & 0.637 \\
            & $p=0.95$ & 0.017 & 0.213 & 0.655 & 0.009 & 0.155 & 0.624 & 0.008 & 0.149 & 0.624 \\
            & $p=1.0$ & \bf{0.025} & \bf{0.277} & 0.633 & \bf{0.016} & \bf{0.229} & 0.603 & \bf{0.016} & \bf{0.229} & 0.603 \\
        \bottomrule
    \end{tabular}}
    \caption{Automatic diversity evaluations across models and decoding methods for each response length. The decoding methods represent a parameter sweep over the $p$ value in nucleus sampling, where $p=1$ corresponds to completely random sampling. The fusion model is a baseline from \citet{fan_hierarchical_2018}.}
    \label{tab:auto-results-appendix}
\end{table*}

\end{document}